\documentclass{article}
\usepackage{amssymb}
\usepackage{booktabs}
\usepackage{wrapfig}
\usepackage{amsmath}
\usepackage{graphicx}
\usepackage{enumitem}
\usepackage{makecell}
\usepackage{multirow}
\usepackage[table]{xcolor}
\usepackage{tocloft}
\usepackage{caption}
\usepackage{booktabs}
\usepackage{xcolor}
\usepackage{booktabs}
\usepackage{pifont}
\newcommand{\cmark}{\ding{51}} 
\newcommand{\xmark}{\ding{55}}  

\usepackage[preprint]{corl_2025} 

\definecolor{myblue}{RGB}{0,100,200}
\definecolor{mygreen}{HTML}{009900}

\usepackage{xspace}

\title{TrackVLA: Embodied Visual Tracking in the Wild}

\author{
    Shaoan Wang$^{*,1,2}$ \ \
    Jiazhao Zhang$^{*,1,2}$ \ \
    Minghan Li$^2$  \ \
    Jiahang Liu$^2$ \ \
    Anqi Li$^{1,2}$ \\
    \textbf{Kui Wu}$^3$ \ \
    \textbf{Fangwei Zhong}$^4$ \ \
    \textbf{Junzhi Yu}$^1$  \ \
    \textbf{Zhizheng Zhang}$^{\dag,2,5}$ \ \
    \textbf{He Wang}$^{\dag,1,2,5}$
}

\begin{document}
\maketitle
\vspace{-1cm}

\begin{abstract}
    Embodied visual tracking is a fundamental skill in Embodied AI, enabling an agent to follow a specific target in dynamic environments using only egocentric vision. This task is inherently challenging as it requires both accurate target recognition and effective trajectory planning under conditions of severe occlusion and high scene dynamics. Existing approaches typically address this challenge through a modular separation of recognition and planning.
    In this work, we propose TrackVLA, a Vision-Language-Action (VLA) model that learns the synergy between object recognition and trajectory planning. Leveraging a shared LLM backbone, we employ a language modeling head for recognition and an anchor-based diffusion model for trajectory planning. To train TrackVLA, we construct an Embodied Visual Tracking Benchmark (EVT-Bench) and collect diverse difficulty levels of recognition samples, resulting in a dataset of 1.7 million samples.
    Through extensive experiments in both synthetic and real-world environments, TrackVLA demonstrates SOTA performance and strong generalizability. It significantly outperforms existing methods on public benchmarks in a zero-shot manner while remaining robust to high dynamics and occlusion in real-world scenarios at 10 FPS inference speed. Our project page is: \url{https://pku-epic.github.io/TrackVLA-web}.

\end{abstract}


\vspace{-12pt}
\begin{figure}[h]
    \centering
    \includegraphics[width=1\linewidth]{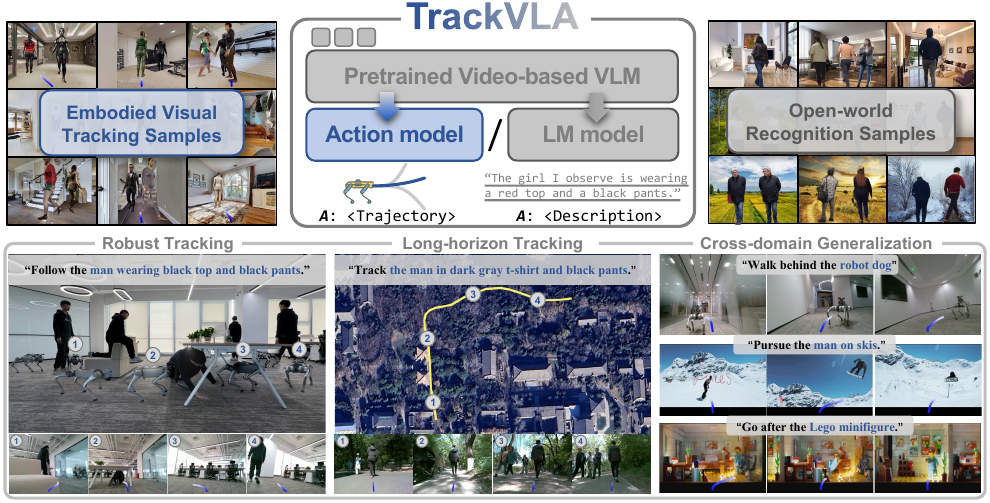}
    \caption{\textbf{TrackVLA} is a vision-language-action model capable of simultaneous object recognition and visual tracking, trained on a dataset of 1.7 million samples. It demonstrates robust tracking, long-horizon tracking, and cross-domain generalization across diverse challenging environments.}
    \label{fig:teaser}
\end{figure}

\vspace{-10pt}
\section{Introduction}
\label{sec:intro}
\vspace{-5pt}

Embodied visual tracking (EVT)~\citep{maalouf2024follow,zhang2018coarse,zhong2019ad,zhong2019ad+,zhong2023rspt,zhong2024empowering} requires the agent to persistently track a given target, which is a fundamental capability of embodied AI~\citep {duan2022survey} and widely demanded in robotics~\citep{mavrogiannis2023core, puig2023habitat}. 
\vspace{-0.6em}
\hrule
\vspace{-0.2em}
\noindent
{\small
$^*$Equal contribution.  (email: \texttt{wangshaoan@stu.pku.edu.cn}, \texttt{zhngjizh@gmail.com}). \\
$^\dag$Corresponding authors. (email: \texttt{zhangzz@galbot.com}, \texttt{hewang@pku.edu.cn}). \\
$^1$Peking University \ $^2$Galbot \ $^3$Beihang University \ $^4$Beijing Normal University \ $^5$BAAI
}
\vspace{1em}

This task is particularly challenging due to its reliance on two tightly coupled skills: (1) Target recognition, the ability to accurately identify and distinguish the target, and (2) Trajectory planning, the capacity to determine optimal actions for effective tracking. The interplay between recognition and planning becomes especially demanding under challenging conditions, such as the presence of severe occlusion and highly dynamic scenes.


Toward achieving robust embodied visual tracking, existing methods~\citep{maalouf2024follow,zhang2018coarse,zhong2019ad,zhong2019ad+,zhong2023rspt,li2020pose} typically address this challenge by decoupling recognition and trajectory planning into a detection model and a planning model, respectively. These approaches benefit from rapid advancements in visual foundation models~\citep{kirillov2023segment, ravi2024sam, liu2023grounding} and policy learning techniques (\textit{e.g.}, imitation learning~\citep{zhang2024uni} and reinforcement learning~\citep{zhong2019ad, luo2018end}). Despite demonstrating early progress, these methods are limited to category-level tracking in relatively open areas. This is because their loosely coupled design causes error accumulation between the recognition model and the planning model—\textit{e.g.}, an incorrect recognition may result in faulty planning, and vice versa. 


To achieve synergy between target recognition and trajectory planning, a versatile model must master both recognition and tracking capabilities. In this work, we propose TrackVLA, a vision-language-action model featuring a unified framework that integrates target recognition and trajectory planning. Specifically, both tasks utilize the same token encoding and LLM forwarding mechanism to predict the next token, while decoding is task-dependent. For the recognition task, TrackVLA employs a language modeling head to decode textual responses. For the planning task, TrackVLA leverages an anchor-based diffusion head to generate waypoint trajectories. Both tasks are trained jointly, optimizing TrackVLA to achieve tight coupling between recognition and planning.




To enable TrackVLA's acquisition of both recognition and planning capabilities, we collect 855K video recognition samples and 855K robot tracking samples. For recognition, we construct a human recognition dataset based on a public ReID dataset~\citep{zuo2024plip} and leverage open-world VQA datasets~\citep{song2024moviechat, chen2024panda, yu2019activitynet}. For embodied visual tracking data, we gather samples from a self-developed embodied visual tracking benchmark (EVT-Bench), which includes over 100 high-fidelity humanoid avatars moving randomly in simulated scenes. Both recognition and tracking samples were collected at varying difficulty levels to enable comprehensive training of TrackVLA.


We conduct extensive experiments on both synthetic and real-world environments, and we find that TrackVLA demonstrates superior performance with strong generalizability. TrackVLA archives SOTA performance in public benchmark Gym-UnrealCV~\citep{qiu2017unrealcv} in a zero-shot manner, and significantly outperforms baselines in self-built benchmark EVT-Bench that involves detailed language input and crowded environments. Furthermore, TrackVLA exhibits exceptional sim-to-real generalization capability, enabling robust tracking of previously unseen objects in novel environments at 10 FPS inference speed. \textit{We will make TrackVLA and EVT-Bench publicly available to benefit the community.}

\vspace{-8pt}
\section{Related Works}
\label{sec:related}
\vspace{-8pt}

\textbf{Embodied Visual Tracking.}
The task requires agents to continuously pursue dynamic targets based on visual observations, relying on accurate target recognition and optimal trajectory planning. In real-world applications, human following~\citep{ye2024person, ye2025rpf, francis2025principles} represents the most extensively studied scenario within this domain. 
While many recent works~\citep{puig2023habitat,luo2018end,luo2019end,devo2021enhancing,zeng2024poliformer,zhong2021towards,bajcsy2024learning,scofano2024following} decouple perception and planning into two separate modules—often incorporating visual foundation models~\citep{kirillov2023segment} to enhance perception and employing reinforcement learning for planning—they frequently suffer from error accumulation due to the separation of detection and planning, as well as low training efficiency. To address this, some approaches leverage offline RL~\citep{zhong2024empowering, shah2022offline} to boost training efficiency.
However, the aforementioned approaches lack support for natural language inputs, which significantly limits their applicability in real-world human-robot interaction scenarios. To address this limitation, Uni-NaVid~\citep{zhang2024uni} introduced a vision-language-action (VLA) model that enables human following via large-scale imitation learning in simulation. Nonetheless, its reliance on a discrete action space hinders adaptability in complex, real-world environments. In contrast, TrackVLA integrates target recognition and trajectory planning into a unified training framework, achieving synergy between robust perception and flexible motion control, and demonstrating superior embodied visual tracking performance in real-world deployments.


\textbf{Embodied Navigation.}
Embodied navigation \citep{Zhang2024VisionandLanguageNT, wu2024embodied, sridhar2024nomad, long2024instructnav, shah2023gnm, zhang2024navid} is a fundamental topic in embodied AI, requiring agents to actively navigate within environments to complete given natural language instructions. Recent advances in embodied navigation have led to the emergence of various subtasks, including Vision-Language Navigation \citep{zhang2024uni, zhou2025navgpt, zhang2024navid}, Object Navigation \citep{kuang2024openfmnav, chaplot2020object,zhang20233d, cao2024cognav}, and Embodied Question Answering \citep{islam2023eqa, majumdar2024openeqa}, among others. However, most current embodied navigation tasks are designed for static indoor environments, overlooking the inherently dynamic nature of real-world environments. In this work, we focus on a challenging embodied navigation task: Embodied Visual Tracking (EVT), which requires identifying a moving target and continuously tracking it in highly dynamic and occluded environments.

\textbf{Vision-Language-Action Models.}
Given the impressive generalization capabilities of Vision-Language Models (VLMs) \citep{team2023gemini, shen2024longvu, steiner2024paligemma, chiang2023vicuna}, Vision-Language-Action (VLA) models have garnered growing attention in the embodied AI community by extending pre-trained VLMs with action generation capabilities. Recently, numerous studies have explored the use of VLA models for tasks such as manipulation \citep{black2024pi_0, intelligence2025pi_, li2024cogact, kim2024openvla, qu2025spatialvla, zhong2025dexgraspvla, ding2024open6dor} and navigation \citep{zhang2024uni, zhang2024navid, cheng2024navila}, demonstrating impressive generalization capabilities. However, most existing VLA models are limited by inference efficiency and have primarily been evaluated in low-dynamic environments. Compared to prior VLA models, TrackVLA exhibits superior performance in highly dynamic environments and demonstrates strong reasoning capabilities for the challenging task of embodied visual tracking.


\begin{figure}[!t]
    \centering
    \includegraphics[width=1\linewidth]{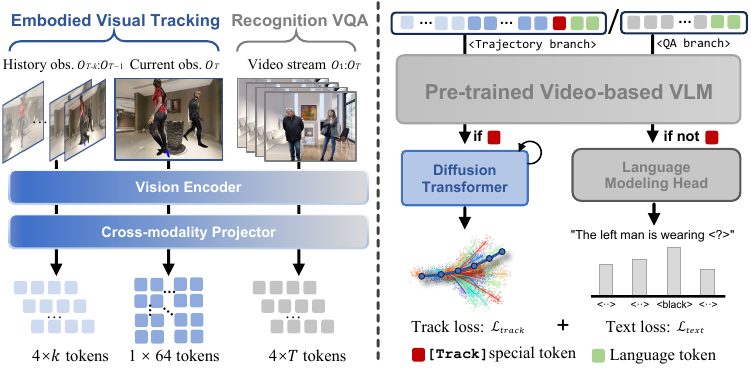}
    \caption{\textbf{Overall pipeline of TrackVLA.} Given a video and a language instruction, TrackVLA outputs either a tracking trajectory for the robot or an answer to the recognition question.}
    \vspace{-1.5em}
    \label{fig:pipeline}
\end{figure}

\vspace{-8pt}
\section{Method}
\label{sec:method}
\vspace{-8pt}

\textbf{Embodied Visual Tracking Formulation}. We formulate embodied visual tracking task as: At each timestamp $T$, given a natural language instruction $\mathcal{I}$, which describes the appearance of a specific target, and an egocentric RGB observation consisting of a sequence of frames~$\mathcal{O}_T = \{\mathbf{x}_1, \cdots, \mathbf{x}_T\} $, the agent is required to output the next action $a_T \in \mathbb{A}=\{v, \omega\}$ to continuously follow the described target in unseen environments, where $\mathbb{A}$ is the action space including linear velocity $v$ and angular velocity $\omega$ of the agent. The task is considered successful if the agent is able to consistently maintain an appropriate following distance (1–3~m) from the target while facing toward it.

\textbf{TrackVLA overview.} As shown in Fig.~\ref{fig:pipeline}, TrackVLA extends video-based VLM/VLA approaches~\cite{li2023llama, zhang2024navid, zhang2024uni} by introducing a parallel prediction branch for both trajectory planning and target recognition. For trajectory planning, TrackVLA organizes online-captured video data, combining historical and current observations, and concatenates them with tracking instructions and a special tracking token. A diffusion transformer then decodes the output tokens from a large language model (implemented with Vicuna-7B~\cite{chiang2023vicuna}) into waypoints. For recognition tasks, all video frames are encoded identically and processed in a conventional autoregressive manner. We present the detailed architecture of TrackVLA in Sec.~\ref{sec:method} and its corresponding dataset in Sec.~\ref{sec:data}.



\subsection{TrackVLA Architecture}
\label{sec:trackvla}

\textbf{Observation Encoding.}
Given the egocentric RGB sequence~$\mathcal{O}_T = \{\mathbf{x}_1, \cdots, \mathbf{x}_T\} $, we employ a pre-trained vision encoder (EVA-CLIP~\citep{sun2023eva}) to extract visual features $\mathbf{V}_{1:T} \in \mathbb{R}^{N \times C}$, where $N$ is the number of patch (set to 256) and $C$ represents the embedding dimension. 
To address this, we apply a grid pooling strategy~\cite{zhang2024navid,zhang2024uni} (Fig.~\ref{fig:pipeline}, left) to the visual features, generating more compact representations. Specifically, we use two resolution scales:
\begin{flalign}
\mathbf{V}^{\textcolor{blue}{\text{fine}} / \textcolor{mygreen}{\text{coarse}}} &= GridPool(\mathbf{V}, \textcolor{blue}{\frac{64}{N}} or \textcolor{mygreen}{\frac{4}{N}})
\end{flalign}
where $V_\text{fine}\in\mathbb{R}^{64 \times C}$ provides fine-grained observations, while $V_\text{coarse}\in\mathbb{R}^{4 \times C}$ offers coarse-grained observations. To optimally balance token length and performance, we empirically use fine-grained features $V_\text{fine}\in\mathbb{R}^{64 \times C}$ for the latest tracking observation to enhance target identification, while coarse-grained tokens are used for historical tracking and VQA-based recognition.


To ensure consistent inference speed during tracking, we employ a sliding window mechanism to retain only the latest $k$ frames (set to 32 in our implementation). For embodied visual tracking, we structure the visual token sequence as: $\mathcal{V}^\text{track}_T = \{\mathbf{V}^\text{coarse}_{T-k},..., \mathbf{V}^\text{coarse}_{T-1}, \mathbf{V}^\text{fine}_{T}\}$, while for the video question answering (VQA) recognition task, we construct the sequence as: $\mathcal{V}^\text{VQA}_T = \{\mathbf{V}^\text{coarse}_{1},..., \mathbf{V}^\text{coarse}_T\}$. Following established Vision-Language Models (VLMs)~\cite{liu2023llava,li2023llama}, we use a cross-modality projector $\mathcal{P}(\cdot)$ (a 2-layer MLP) to project visual features into the latent space of the Large Language Model: $\mathbf{E}^V_T = \mathcal{P}(\mathcal{V}_T)$.

\textbf{Large Language Model Forwarding.}
We concatenate the visual tokens $\mathbf{E}_T^V$ with the language tokens $\mathbf{E}^{I}$ (adding a special \texttt{[Track]} token for the tracking task) and feed them into the LLM  (Fig.~\ref{fig:pipeline} Right) to obtain the predicted token $\mathbf{E}_T^\text{pred}$.
The predicted token is then processed differently depending on the task (determined by the presence of the \texttt{[Track]} token). For recognition tasks, we use the standard language modeling head to decode the token auto-regressively into a vocabulary word~\citep{touvron2023llama}. For tracking tasks, $\mathbf{E}_T^\text{pred}$ serves as conditional input to our action head model, which generates waypoint trajectories for navigation.


\begin{wrapfigure}{r}{0.45\linewidth}
    \centering
    \vspace{-10pt} 
    \includegraphics[width=\linewidth]{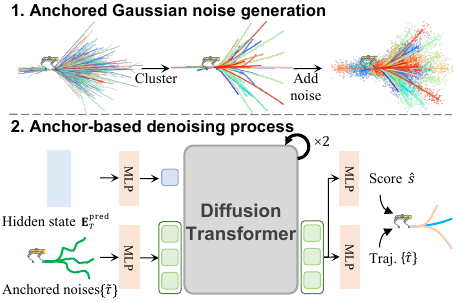}
    \caption{Anchor-based Diffusion Action Model.}
    \vspace{-1em}
    \label{fig:action_model}
\end{wrapfigure}
\textbf{Anchor-based Diffusion Action Model.} We employ an anchor-based diffusion model~\cite{liao2024diffusiondrive} that performs denoising from predefined anchors to generate waypoints. These predefined anchors provide initial coarse trajectories that significantly reduce the required denoising iterations, yielding a 5× speedup compared to vanilla diffusion policies~\citep{sridhar2024nomad,chi2023diffusion}.
As shown in Fig.~\ref{fig:action_model}, we first collect all trajectories from the training data and apply K-means clustering~\citep{arthur2006k} to obtain a set of trajectory anchors $\{\tau_i\}_{i=1}^{M}$, where $M$ denotes the number of anchors. Each anchor $\tau_i = {(x_i, y_i, \theta_i)}_{i=1}^{N_w}$ represents a robot trajectory pattern, where $N_w$ is the number of waypoints in each trajectory.
We then perturb each anchor with Gaussian noise to create noised anchors $\{\tilde{\mathbf{\tau}_i}\}_{i=1}^M$.
Our action model $\mathcal{A}_{\theta}(\cdot)$ takes the set of noised anchors $\{\tilde{\mathbf{\tau}_i}\}_{i=1}^M$ and the condition $\mathbf{E}_T^\text{pred}$ as input, and outputs: the denoised trajectories $\{\hat{\mathbf{\tau}}_i\}_{i=1}^{M}$ and the corresponding trajectory classification scores $\{\hat{s}_i\}_{i=1}^{M}$:
\begin{equation}
\left\{ \hat{s}_i, \hat{\mathbf{\tau}}_i \right\}_{i=1}^{M} = \mathcal{A}_\theta\left( \left\{ \tilde{\mathbf{\tau}}_i \right\}_{i=1}^{M}, \mathbf{E}_T^\text{pred} \right)
\end{equation}

For each sample, we label the anchor trajectory closest to the ground truth $\tau_{gt}$ as positive (\mbox{$s_{\text{nearest}}=1$}), and all others as negative ($s_{\text{else}}=0$).
We then jointly optimize the trajectory regression loss and the score prediction loss. The tracking loss $\mathcal{L}_{track}$ is defined as:
\begin{equation}
    \mathcal{L}_\text{track}=\sum_{i=1}^{M}[s_iMSE(\hat{\tau}_i,\tau_{gt})+\lambda BCE(\hat{s}_i,s_i)]
\end{equation}
where $\lambda$ is a balancing parameter. Here, we adopt the Diffusion Transformer (DiT)~\citep{peebles2023scalable} for denoising and the anchor-based diffusion policy only needs two denoising steps. Given a batch of input sequences, the overall training loss $\mathcal{L}$ is defined as a weighted combination of the tracking loss $\mathcal{L}_{track}$ and the text prediction loss $\mathcal{L}_{text}$, formulated as $\mathcal{L} = \mathcal{L}_{track} + \alpha \mathcal{L}_{text}$, where $\alpha$ is also a balancing parameter. More details can be found in the Appendix.


\subsection{Implementation Details}
\textbf{Training Details.} During training, we follow the standard practice in vision-language modeling (VLM)~\cite{liu2023llava} by training for only one epoch. Additionally, we freeze the parameters of the vision encoder throughout training. 
\textbf{Inference Details.} During inference, we use a special token \texttt{[Track]} to indicate the current task. When the \texttt{[Track]} token is present, LLM performs only a single-step autoregression and passes the output hidden state to the action model to predict trajectories. We apply the DDIM~\citep{song2020denoising} update rule for denoising with only two steps, and select the trajectory $\hat{\tau}_k$ corresponding to the top-1 score $\hat{s}_k$ as the final output. Otherwise, LLM conducts full autoregressive decoding to answer the given question based on visual observations. More details are provided in the Appendix.

\vspace{-8pt}
\section{Data Collection }
\label{sec:data}
\vspace{-8pt}

To train our parallel branch TrackVLA, we collect both embodied visual tracking data (855K samples) and video-based question-answering data (855K samples), where we empirically find that a 1:1 ratio yields the best performance (Fig.~\ref{fig:ratio}).
For tracking samples (Sec.~\ref{sec:tracking_data}), we develop a custom avatar-following simulator and collect a diverse dataset spanning challenging scenarios.
For recognition samples (Sec.~\ref{sec:vqa_data}), we construct a video question-answering dataset that requires the agent to describe or distinguish target objects amidst complex backgrounds and distractors.

\subsection{Embodied Visual Tracking Data}
\label{sec:tracking_data}

\begin{figure}[!t]
    \centering
    \includegraphics[width=1\linewidth]{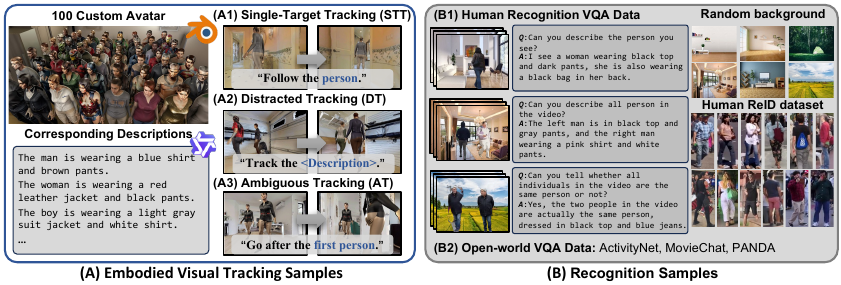}
    \caption{\textbf{Overview of the training datasets used in TrackVLA.} We collect 855K embodied visual tracking samples and 855K open-world recognition samples to jointly enhance the robust recognition and tracking capabilities of TrackVLA.}
    \vspace{-1em}
    \label{fig:dataset}
\end{figure}

\textbf{Embodied Visual Tracking Simulator.} We build our embodied visual tracking simulator based on Habitat 3.0~\citep{puig2023habitat}, which provides an off-the-shelf simulation engine for collision detection and rendering. Our main enhancements include two aspects: (1) \textbf{\textit{Humanoid Avatar Generation.}} We implement a fully automated pipeline for generating and annotating diverse humanoid avatars (Fig.~\ref{fig:dataset} (A)). Specifically, we adopt the SMPL-X human model and initialize the avatars with random shapes and randomly sampled UV texture maps (ATLAS dataset~\citep{liu2024texdreamer}). We then use a vision-language model (Qwen-VL2.5~\cite{bai2025qwen2}) to obtain corresponding textual descriptions of the avatars. (2) \textbf{\textit{Natural Human Behaviors.}} We assign each avatar a series of targets that it must reach in order, with on-and-off walking states. The walking speed is randomly sampled from a natural human walking speed range of [1.0 m/s - 1.5 m/s]~\citep{knoblauch1996field}. Furthermore, we employ the ORCA algorithm~\citep{van2011reciprocal} to enable dynamic collision avoidance and responsive interactions, resulting in more natural behavior. For more details, please refer to the Appendix.





\textbf{Embodied Visual Tracking Benchmark.}
Based on our simulator, we construct the Embodied Visual Tracking Benchmark (EVT-Bench) to comprehensively evaluate embodied visual tracking capabilities. We generate 100 diverse humanoid avatars and corresponding descriptions and utilize 804 scene environments from HM3D~\citep{ramakrishnan2021hm3d} and MP3D~\citep{chang2017matterport3d}. A total of 25,986 episodes are generated and subsequently divided into training and testing sets, ensuring no overlap of avatars or scenes between the two splits. The training set consists of 21,771 episodes across 703 scenes, while the testing set includes 4,215 episodes across 101 unseen scenes. To comprehensively evaluate algorithm performance across different scenarios, EVT-Bench is divided into three sub-task categories of increasing difficulty. Each sub-task contains 7,257 episodes for training and 1,405 episodes for testing. We list each category of tracking task below:
\vspace{-7pt}
\begin{itemize}[leftmargin=2em]
    \setlength\itemsep{-0em}
    \item \textit{\textbf{Single-Target Tracking (STT)}} evaluates the model’s basic following ability with simple instructions like \textit{“Follow the person/man/woman"}. 
    \item \textit{\textbf{Distracted Tracking (DT)}} evaluates the model's recognition abilities with fine-grained descriptions of the target, such as \textit{“Follow the light-skinned man in a black suit with a white belt”}. 
    \item \textit{\textbf{Ambiguity Tracking (AT)}} evaluates the model’s ability to identify the correct target when distractors with identical appearances are present. The instructions are intentionally ambiguous, such as \textit{“Follow the first person you see”}.
\end{itemize}
\vspace{-7pt}


\textbf{Tracking Data Collection.}
We collect 885K embodied visual tracking samples in the EVT-Bench training split, covering three sub-tasks of varying difficulty. Each sample includes a navigation history (RGB sequence), a target description, and the corresponding expert trajectory $\tau_{gt}$. Additional details regarding the benchmark and data collection are provided in the Appendix. The EVT-Bench will be made publicly available to benefit the research community.

\subsection{Video Question Answering Dataset}
\label{sec:vqa_data}
Despite our considerable efforts to incorporate diverse avatars and indoor scenes, the tracking samples remain limited to synthetic environments. To equip TrackVLA with open-world recognition capabilities (beyond tracking samples), we further collect a total of 855K recognition samples and jointly train them with the tracking samples. Specifically, the recognition video question-answering (VQA) samples consist of 362K human recognition samples and 493K open-world VQA samples.



For the human recognition VQA data, we leverage SYNTH-PEDES~\citep{zuo2024plip}, a large-scale person-text dataset, to construct VQA samples that require TrackVLA to identify or describe individuals in videos featuring randomly composed human subjects and background scenes. Each sample is created by placing 1–3 randomly selected human images onto diverse backgrounds, with accompanying textual descriptions detailing each individual’s attributes, their relative spatial positions, and whether they represent the same identity.
In addition to human recognition samples, we also incorporate publicly available VQA samples \citep{chen2024panda, song2024moviechat, yu2019activitynet} that provide open-world captions. These samples enhance TrackVLA’s ability to recognize open-world targets. (see Table~\ref{tab:evt_bench}). 



\vspace{-8pt}
\section{Experiments}
\label{sec:exp}
\vspace{-8pt}

We conduct experiments to evaluate TrackVLA from three perspectives: (1) How well does TrackVLA perform in embodied visual tracking? (2) How strong is its target recognition ability? (3) The effectiveness of proposed designs.

\subsection{Experiment Setups}

\textbf{Benchmarks.}
We evaluate our method on a public benchmark Gym-UnrealCV~\citep{qiu2017unrealcv} (zero-shot evaluation) and our proposed benchmark EVT-Bench.
\textbf{Baselines.}
We conduct a comprehensive comparison of our proposed approach against current state-of-the-art models, which can be categorized into three groups: (1) model-based method IBVS~\citep{gupta2016novel}, (2) reinforcement learning (RL)-based methods including DiMP~\citep{bhat2019learning}, SARL~\citep{luo2019end}, AD-VAT~\citep{zhong2019ad}, AD-VAT+\citep{zhong2019ad+}, TS~\citep{zhong2021towards}, EVT~\citep{zhong2024empowering}, and PoliFormer~\citep{zeng2024poliformer}, and (3) imitation learning (IL)-based method Uni-NaVid~\citep{zhang2024uni}.
\textbf{Metrics.}
To evaluate tracking performance, we use the standard evaluation metrics from Gym-UnrealCV~\citep{qiu2017unrealcv} and EVT-Bench, including success rate (SR), average episode length (EL), tracking rate (TR), and collision rate (CR). Further details of the experiment setup are provided in the Appendix.

\setlength{\tabcolsep}{3pt}
\begin{table}[t]
    \centering
    \begin{minipage}[t]{0.42\textwidth}
        \centering
        \resizebox{\textwidth}{!}{
        \begin{tabular}{lccc}
            \toprule
            \multirow{3}{*}{Methods} 
            & \textbf{Single} & \textbf{Distractor} & \textbf{Unseen} \\
            & \textbf{Target} &                      & \textbf{Objects} \\
            & EL$\uparrow$ / SR$\uparrow$  & EL$\uparrow$ / SR$\uparrow$  & EL$\uparrow$ / SR$\uparrow$ \\
                    \midrule
            DiMP~\citep{bhat2019learning} & 367/0.58 & 309/0.27 & -/- \\
            SARL~\citep{luo2019end} & 394/0.57 & 240/0.14 & -/- \\
            AD-VAT~\citep{zhong2019ad} & 416/0.62 & 220/0.12 & -/- \\
            AD-VAT+~\citep{zhong2019ad+} & 454/0.76 & 224/0.12 & -/- \\
            TS~\citep{zhong2021towards} & 474/0.86 & 371/0.48 & -/-\\
            EVT~\citep{zhong2024empowering} & 490/0.95 & 459/0.81 & 480/0.96 \\
            \rowcolor{gray!20}
            Ours & \textbf{500}/\textbf{1.00} & \textbf{474}/\textbf{0.91} &  \textbf{500}/\textbf{1.00} \\
            \bottomrule
        \end{tabular}
        }
        \caption{Zero-shot performance on Gym-UnrealCV.}
        \label{tab:evt_bench}
    \end{minipage}
    \hfill
    \begin{minipage}[t]{0.57\textwidth}
        \centering
        \resizebox{\textwidth}{!}{
        \begin{tabular}{lccc}
            \toprule
            \multirow{2}{*}{Methods}& \textbf{\textit{STT}} & \textbf{\textit{DT}} & \textbf{\textit{AT}}  \\
            & SR$\uparrow$/ TR$\uparrow$ / CR$\downarrow$ & SR$\uparrow$/ TR$\uparrow$ / CR$\downarrow$ & SR$\uparrow$/ TR$\uparrow$ / CR$\downarrow$ \\
            \midrule
            IBVS$\dag~\citep{gupta2016novel} $ & 42.9/56.2/3.75 & 10.6/28.4/6.14 & 15.2/39.5/\textbf{4.90} \\
            PoliFormer$\dag~\citep{zeng2024poliformer} $ & 4.67/15.5/40.1 & 2.62/13.2/44.5 & 3.04/15.4/41.5 \\
            EVT~\citep{zhong2024empowering} & 24.4/39.1/42.5 & 3.23/11.2/47.9 & 17.4/21.1/45.6\\
            EVT$\ddag$~\citep{zhong2024empowering} & 32.5/49.9/40.5 & 15.7/35.7/53.3 & 18.3/21.0/44.9 \\
            Uni-NaVid~\citep{zhang2024uni} & 25.7/39.5/41.9 & 11.3/27.4/43.5 & 8.26/28.6/43.7 \\
            \rowcolor{gray!20}
            Ours & \textbf{85.1}/\textbf{78.6}/\textbf{1.65} & \textbf{57.6}/\textbf{63.2}/\textbf{5.80} & \textbf{50.2}/\textbf{63.7}/17.1 \\
            \bottomrule
        \end{tabular}
        }
        \caption{Performance on EVT-Bench. $\dag$: Use GroundingDINO~\citep{liu2023grounding} as the open-vocabulary detector. $^\ddag$: Use SoM~\citep{yang2023set}+GPT-4o~\citep{openai2024introducing} as the visual foundation model.}
        \label{tab:evt-bench}
    \end{minipage}
    
\vspace{-20pt}
\end{table}
\subsection{Quantitative Comparison}

\textbf{Zero-shot performance on Gym-UnrealCV}. We first evaluate our method on a public tracking benchmark, Gym-UnrealCV, in a zero-shot manner. The results can be found in Table~\ref{tab:evt_bench}, where our method significantly outperforms existing baselines. Particularly in the \texttt{Single Target} and \texttt{Unseen Objects} tasks, our method successfully tracks the target throughout the entire tasks (500 steps over 100 episodes). For the more challenging \texttt{Distractor} task, where the agent must identify and track the initially seen target among identical distractors, our method still surpasses the previous state-of-the-art EVT~\cite{zhong2024empowering} (with EL $3.25\%\uparrow$ and SR $12.3\%\uparrow$). The zero-shot performance of our method clearly demonstrates its generalization capability in tracking and recognition performance.


\textbf{Performance on EVT-Bench}. We further evaluate our method on the proposed benchmark, EVT-Bench, as shown in Table~\ref{tab:evt-bench}. TrackVLA significantly outperforms existing approaches across all three tasks (\texttt{STT}, \texttt{DT}, and \texttt{AT}), demonstrating its robust and comprehensive tracking capabilities, particularly in comparison to the VLA method Uni-NaVid~\cite{zhang2024uni}. However, despite these improvements, we observe a noticeable performance drop when transitioning from single-target tracking (STT) to distracted tracking (DT) and ambiguity tracking (AT), highlighting the challenges of accurately recognizing and following a specified target in complex environments with distractors. We believe our benchmark can benefit the research community by providing a well-defined target for future studies. 

\begin{wraptable}{r}{0.4\textwidth}
    \centering
    \vspace{-10pt}
    \resizebox{0.4\textwidth}{!}{
    \begin{tabular}{lccc}
        \toprule
        Methods & ACC$\uparrow$ &  FPS$\uparrow$ \\
        \midrule
        RexSeek~\citep{jiang2025referring} & 54.3 & 1.1 \\
        LISA++~\citep{yang2023lisa++} & 78.2 & 0.6 \\
        SoM~\citep{yang2023set}+GPT-4o~\citep{openai2024introducing} & \textbf{82.4} & 0.1 \\
        \rowcolor{gray!20}
        Ours w/o VQA & 62.3  & \textbf{10} \\
        \rowcolor{gray!20}
        Ours & 80.7  & \textbf{10} \\
        \bottomrule
    \end{tabular}
    }
    \vspace{-8pt} 
    \caption{Comparison of different methods on recognition ability.}
    \label{tab:passive_bench}
    \vspace{-10pt}
\end{wraptable}
\textbf{Performance on Visual Recognition}. Furthermore, we evaluate the recognition capability of TrackVLA and state-of-the-art VLMs~\citep{jiang2025referring, yang2023lisa++, openai2024introducing} on a recognition task that distinguishes between two randomly selected \textit{unseen} human images from SYNTH-PEDES. The results are presented in Table~\ref{tab:passive_bench}, where we report the accuracy over 2000 samples and the corresponding inference FPS. We find that our method achieves comparable performance to a strong baseline, SoM~\citep{yang2023set} + GPT-4o~\citep{openai2024introducing}, while achieving a 10 FPS inference speed, approximately 100$\times$ faster than GPT-based baselines. Moreover, we observe that co-tuning with VQA samples leads to significant improvements ($29.53\%\uparrow$ in ACC), demonstrating the effectiveness of our recognition-focused VQA samples.


\subsection{Qualitative Results in Real-World}

\begin{figure}[!t]
    \centering
    \includegraphics[width=1\linewidth]{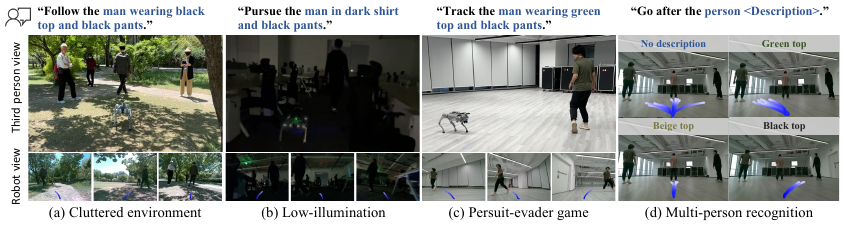}
    \caption{\textbf{Real-world qualitative results of TrackVLA.} TrackVLA is deployed in a zero-shot manner across diverse environments, executing diverse tracking instructions in challenging scenarios.}
    \vspace{-1em}
    \label{fig:gallery}
\end{figure}

We provide qualitative real-world results in Fig.~\ref{fig:gallery}, where we evaluate our method in challenging scenarios, including: (A) cluttered environments, (B) low-lighting conditions, (C) pursuit-evasion tasks, and (D) multi-person recognition. The experimental results demonstrate that TrackVLA exhibits strong sim-to-real transfer capabilities in both recognition and tracking, while maintaining high-frequency inference in real-world scenarios, thereby enabling zero-shot deployment in highly dynamic environments. For additional real-world performance demonstrations, we refer the audience to our supplementary video.


\begin{wrapfigure}{r}{0.4\linewidth}
    \centering
    \vspace{-20pt} 
    \includegraphics[width=1\linewidth]{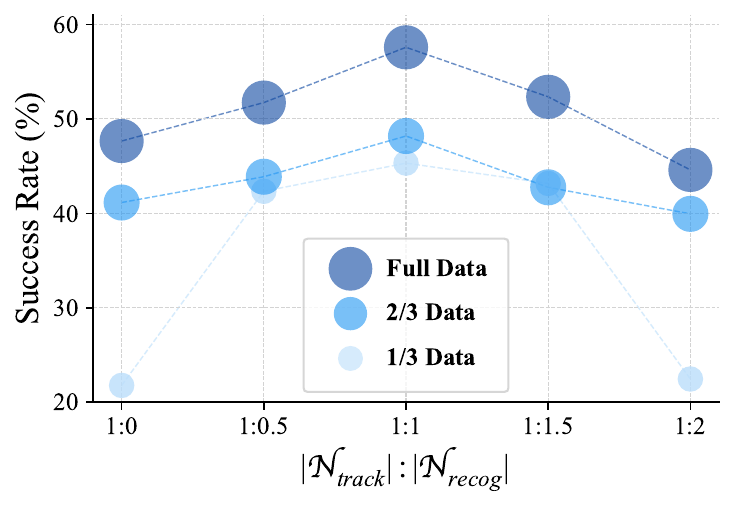}
    \vspace{-15pt}
    \caption{Comparison of different data scales and ratios.}
    \label{fig:ratio}
    \vspace{-20pt}
\end{wrapfigure}

\subsection{Ablation Study}

\textbf{Data Scale and Data Ratio.} We conduct an ablation study on the \texttt{DT} task in EVT-Bench to investigate the influence of training sample scale and ratio. Here, we denote the number of embodied visual tracking samples and open-world recognition samples as $|\mathcal{N}_{track}|$ and $|\mathcal{N}_{recog}|$, respectively. The results are shown in Fig.~\ref{fig:ratio}, where we observe that increasing the scale of training samples consistently improves performance across all data ratios, aligning with the data scaling law. Furthermore, we find that a 1:1 ratio yields the best performance, which may be attributed to more balanced gradient updates~\citep{misra2016cross}.


\begin{wraptable}{r}{0.5\textwidth}
    \centering
    \vspace{-10pt}
    \resizebox{0.5\textwidth}{!}{
        \begin{tabular}{lccccc}
        \toprule
        Model & Params. & SR$\uparrow$ & TR$\uparrow$ & CR$\downarrow$ & time(ms) $\downarrow$ \\
        \midrule
        Autoregressive  & 131M  & 42.6  &  56.9    &  11.7    &  460  \\
        MLP (3-Layers)  & 7M  & 45.8  &  59.9    &  10.1    &   \textbf{0.5}   \\
        MLP (6-Layers)  & 89M  & 52.7  &  61.9    &  9.42   &   0.8  \\
        DP-Base & 89M &  17.9  &  33.8    &  27.7    &   65   \\
        Ours-Small &  13M    & 49.8 &   60.2   &  6.67   &  8    \\
        \rowcolor{gray!20}
        Ours-Base & 89M    & \textbf{57.6} &  \textbf{63.2}   &  \textbf{5.80}    & 13    \\
        \bottomrule
        \end{tabular}}
    \vspace{-8pt}    
    \caption{Comparison of different action models.}
    \vspace{-10pt}
    \label{tab:ablation_action}
\end{wraptable}

\textbf{Action Model Architecture.}
We further evaluate the performance of widely used action models on the \texttt{DT} task in EVT-Bench. As shown in Table~\ref{tab:ablation_action}, our anchor-based diffusion model (Sec.~\ref{sec:trackvla}) outperforms all existing baselines—including Autoregressive, MLP, and vanilla Diffusion Policy (DP)—across all metrics while maintaining high efficiency. Furthermore, we observe that scaling up the DiT backbone in the action model consistently improves performance, suggesting a promising scaling behavior of the action model with diffusion transformers. Additional baseline configurations and experimental details are provided in the Appendix.


\vspace{-8pt}
\section{Conclusions}
\label{sec:conclusion}
\vspace{-8pt}
In this work, we propose TrackVLA, a Vision-Language-Action (VLA) model designed for the embodied visual tracking task. TrackVLA supports the output of both tracking trajectories and text-based responses. It is jointly trained on both embodied visual tracking data and open-world recognition data, enabling it to learn the synergy between these two modalities. To support this, we collect a large-scale dataset consisting of 855K embodied visual tracking samples and 855K open-world recognition samples. Extensive experiments demonstrate its state-of-the-art performance in simulation and strong generalization, enabling zero-shot deployment in real-world scenarios.

\section{Limitations}
\vspace{-8pt}
\label{sec:limitations}
While TrackVLA demonstrates strong performance and efficiency, several limitations remain:
First, the current method relies solely on egocentric observation, limiting TrackVLA to a narrow field of view (typically 90° FOV). Integrating panoramic~\citep{zheng2024towards} or multi-view inputs~\citep{lu2025multi} could mitigate this issue and enhance tracking robustness.
Second, the current approach only employs a waypoint controller and lacks a more flexible local motion controller~\citep{roth2024viplanner}. Incorporating such a controller could improve movement speed and expand reachable areas. We plan to integrate locomotion capabilities into TrackVLA in future work.





\bibliography{reference}  

\appendix
\label{sec:appendix}

\section{Training Details}
Similar to conventional vision-language models (VLMs), TrackVLA follows a two-stage training pipeline. In the first stage, we train the projector of the visual encoder using a large amount of image-caption data~\citep{liu2023llava} to align the visual embedding space with the LLM's latent space. In the second stage, we jointly train the visual projector, the large language model, and the action model using a mixture of the training data. During training, we truncate the diffusion schedule of the action model to at most 50 out of a total of 1000 steps to diffuse the trajectory anchors, which introduces only a small amount of noise. 

TrackVLA is trained on a cluster server equipped with 24 NVIDIA H100 GPUs for approximately 15 hours, totaling 360 GPU hours. The vision encoder (EVA-CLIP~\citep{sun2023eva}) and the large language model (Vicuna-7B~\citep{chiang2023vicuna}) are initialized with their respective pretrained weights, and the vision encoder remains frozen throughout the entire training process. Following standard VLM practices, we train the model for only one epoch. The training is conducted with a learning rate of 2e-5, a total batch size of 196, and a cosine learning rate schedule with linear warm-up. We use the AdamW optimizer for optimization. See Table~\ref{tab:notation} for detailed parameter settings.

\section{Inference Details}
During inference, each input frame is resized to 224$\times$224 and fed into the vision encoder. After obtaining visual tokens, we organize the tokens according to the task type. For the embodied visual tracking task, we prepend a special \texttt{[Track]} token before the instruction tokens and perform only a single-step autoregression with the LLM. The final-layer hidden state output from the LLM is then passed to the action model. We apply 10 out of 1000 diffusion steps to the trajectory anchors and use DDIM to perform 2 denoising steps, resulting in a set of predicted trajectories and corresponding score vectors. The trajectory corresponding to the anchor with the top~1 score is selected as the final output. For the VQA task, we follow the standard autoregressive decoding process of the LLM, and the language modeling head detokenizes the predicted tokens into textual answers. See Table~\ref{tab:notation} for detailed parameter settings.

\begin{table}[h]
\centering
\begin{tabular}{lcc}
\toprule
\textbf{Notation} & \textbf{Shape \& Params.} & \textbf{Description} \\
    \midrule
        $lr$ & 2e-5 & learning rate \\
        $B$ & 196 & batch size \\
        $T$ & 1000 & total diffusion steps \\
        $T_{train}$ & 50 & number of noise addition steps during training \\
        $T_{infer}$ & 10 & number of noise addition steps during training \\
        $N_{step}$ & 2 & denoising steps during inference \\
        $\mathbf{X}$ & 224$\times$224 & input observation size \\
        $N$ & 256 & the number of image patch \\
        $C$ & 1408 & embedding dimension of visual feature \\
        $\alpha$ & 1 & balancing parameter 1 \\
        $\lambda$ & 100 & balancing parameter 2 \\
        $M$ & 40 & the number of trajectory anchors \\
        $N_w$ & 10 & the number of waypoints \\
    \bottomrule
\end{tabular}
\caption{Hyperparameters and notation used in our model.}
\label{tab:notation}
\end{table}

\section{EVT-Bench}
\subsection{Episode Generation}
For each episode, we first sample a motion trajectory for the target humanoid avatar within the navigable area. Each trajectory consists of a start point, a random number of intermediate waypoints (0–2), and an end point. The distance between any two consecutive waypoints must exceed a predefined minimum threshold $d_{min} = 3$ m. After generating the trajectory for the target avatar, the agent is placed near the target’s starting point, with its initial orientation roughly facing the target but perturbed by a random offset within $±30^{\circ}$. For the \textit{DT} and \textit{AT} tasks, distractors are initialized near the target’s trajectory, and their paths are designed to intersect with the target’s trajectory as much as possible to enhance the level of distraction.

\subsection{Evaluation}
In each episode, the target humanoid and distractors move along their predefined trajectories. The evaluated algorithm receives the agent’s observation at each time step and performs inference to generate the corresponding control command, consisting of linear and angular velocities of the agent. The agent then moves according to the speed command. The episode terminates when the target humanoid reaches its destination or when the agent collides with the humanoid.

\subsection{Metric Definitions}
\begin{itemize}[leftmargin=2em]
    \item \textbf{Success Rate (SR)}: This metric evaluates the agent’s tracking ability. An episode is considered successful if, by its end, the agent remains oriented toward the target and maintains a safe distance of 1–3 meters. The success rate is defined as the proportion of successful episodes over the total number of episodes.  
    \item \textbf{Tracking Rate (TR)}: This metric evaluates the tracking quality of the aggent. It is defined as the proportion of steps $S$ where the agent successfully tracks the target to the total number of steps $L$, i.e., $\text{TR} = S/L$.  
    \item \textbf{Collision Rate (CR)}: This metric evaluates the safety of the agent. It is defined as the proportion of episodes that terminate due to a collision between the agent and the target humanoid avatar.
\end{itemize}

\subsection{Humanoid Avatar Gallery}
\begin{figure}[h]
    \centering
    \includegraphics[width=1\linewidth]{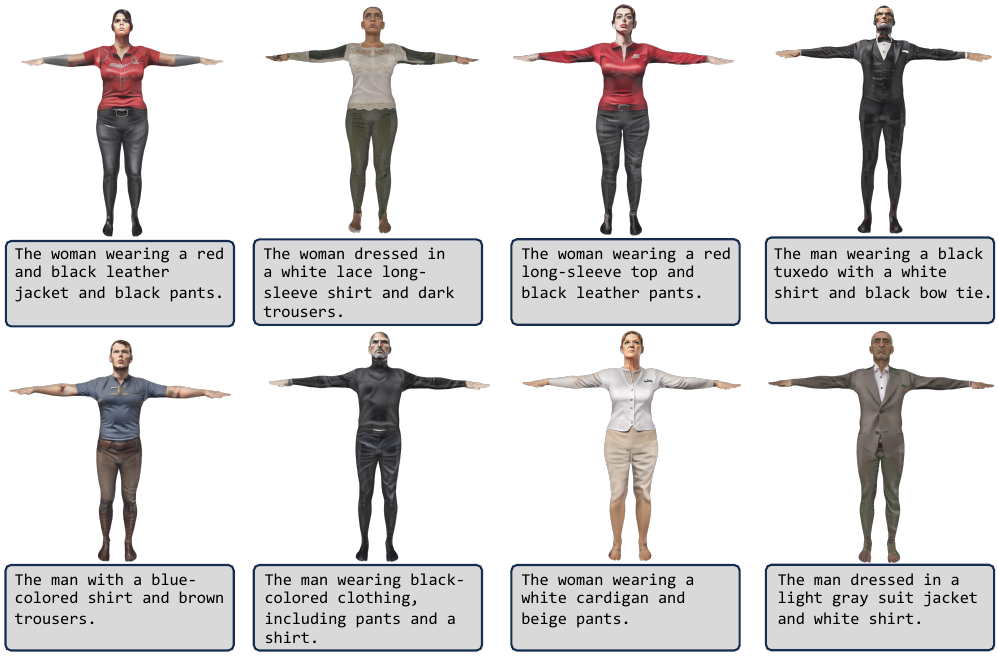}
    \caption{Visualization of the custom humanoid avatars with captions.}
    \label{fig:avatar}
\end{figure}

We provide visualizations of several custom-designed humanoid avatars paired with their descriptive captions, as shown in Fig.~\ref{fig:avatar}

\subsection{Visualization of Training Data}
\begin{figure}[h]
    \centering
    \includegraphics[width=1\linewidth]{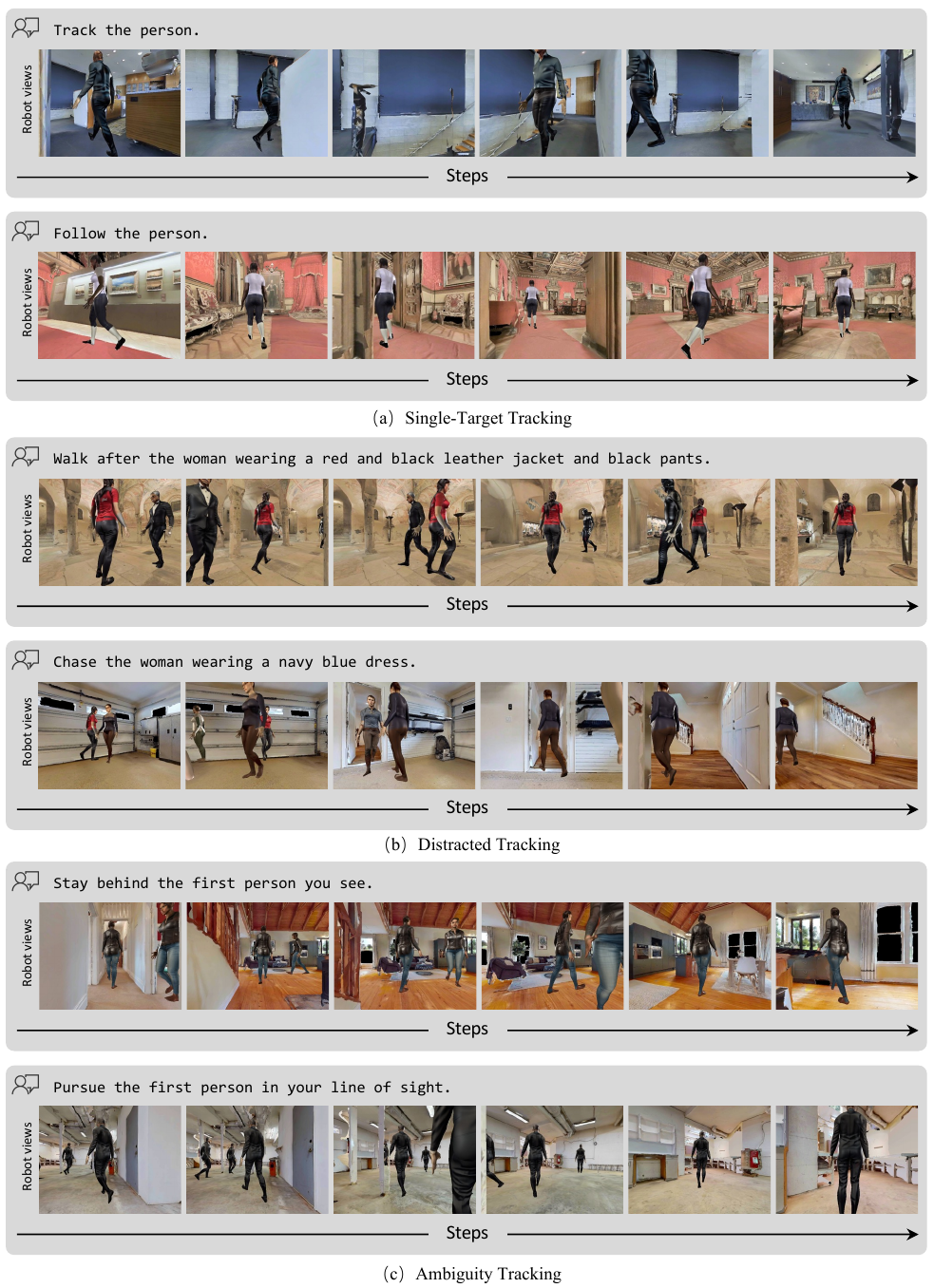}
    \caption{Visualization of the training set of EVT-Bench.}
    \label{fig:data_gallery}
\end{figure}

We provide visualizations of the training set of self-built EVT-Bench, as shown in Fig.~\ref{fig:data_gallery}

\subsection{Qualitative Results on EVT-Bench}
\begin{figure}[h]
    \centering
    \includegraphics[width=1\linewidth]{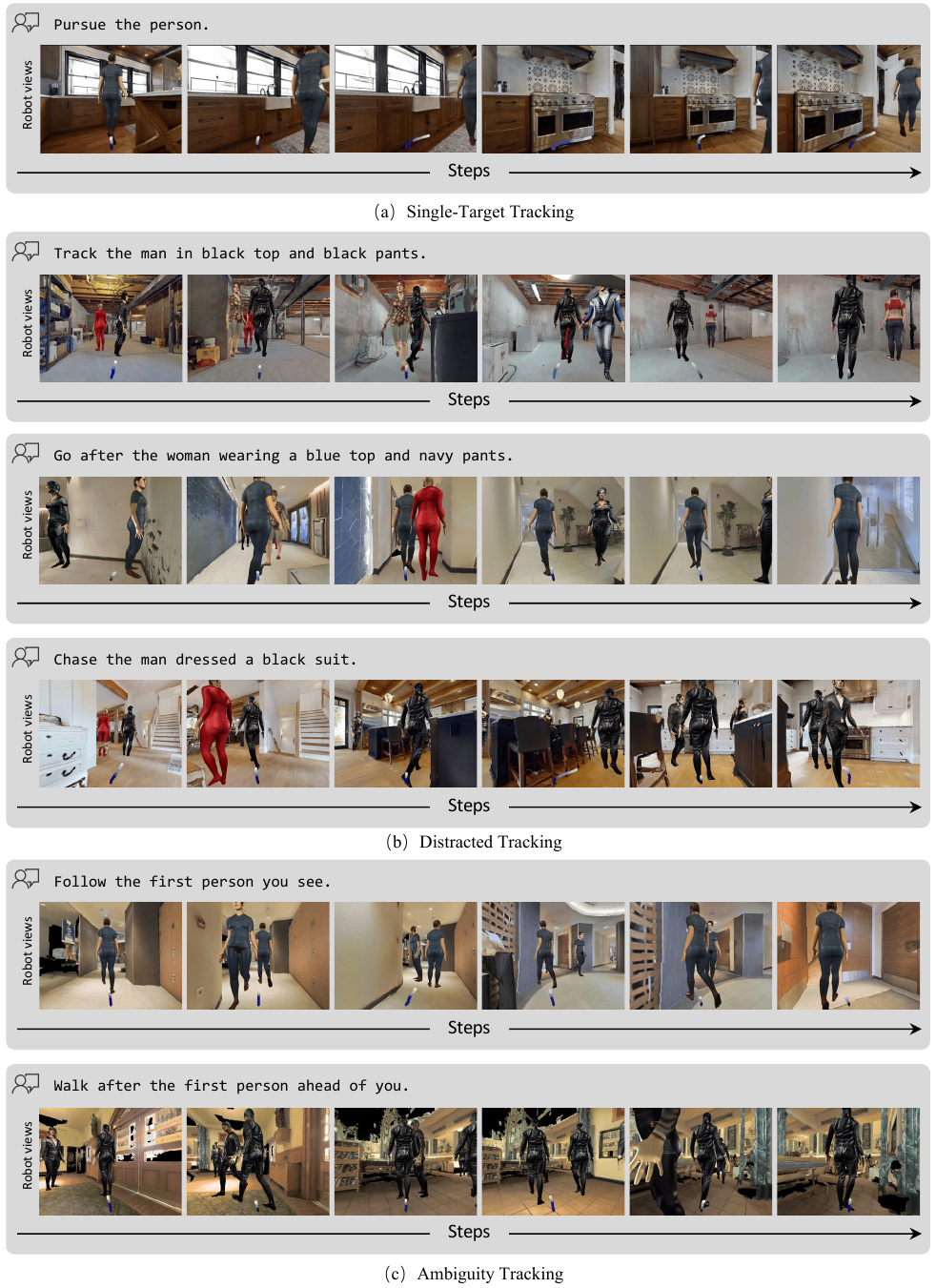}
    \caption{Visualization of TrackVLA on EVT-Bench.}
    \label{fig:evt_infer}
\end{figure}

We provide visual results of TrackVLA on EVT-Bench, shown in Fig.~\ref{fig:evt_infer}.


\section{Detailed Experiments of Gym-UnrealCV}
\subsection{Evaluation}
The evaluation setting of Gym-UnrealCV follows~\citep{zhong2024empowering}, where each episode has a maximum length of 500 steps. The agent’s tracking region is defined as a 90-degree fan-shaped sector with a radius of 750 cm. Success is achieved if the agent keeps the target within this region for the entire episode. A failure occurs if the target remains outside the region for more than 50 consecutive steps.

\subsection{Metric Definitions}
\begin{itemize}[leftmargin=2em]
    \item \textbf{Episode Length (EL)}: The average number of steps per episode over 100 episodes, reflecting the model's long-term tracking capability under predefined termination conditions.
    \item \textbf{Success Rate (SR)}: The percentage of successful episodes out of the total 100 episodes, measuring the model’s robustness in active visual tracking.
\end{itemize}

\subsection{Testing Scenes}
\begin{itemize}[leftmargin=2em]
    \item \textbf{SimpleRoom}: A basic environment designed to verify the model’s fundamental tracking capability.
    \item \textbf{Parking Lot}: An environment featuring occlusions and low-light conditions.
    \item \textbf{UrbanCity}: A typical urban street scene with reflective road surfaces.
    \item \textbf{UrbanRoad}: Similar to UrbanCity with fewer obstacles.
    \item \textbf{Snow Village}: A challenging terrain with uneven surfaces and complex backlighting.
\end{itemize}

\subsection{Baselines}
\begin{itemize}[leftmargin=2em]
    \item \textbf{DiMP}~\cite{bhat2019learning}: Utilizes a pre-trained video tracker to generate target bounding boxes as scene representations and applies a PID controller for motion control.
    
    \item \textbf{SARL}~\cite{luo2019end}: An online reinforcement learning (RL) approach that encodes RGB observations into latent visual features and trains an end-to-end Conv-LSTM policy via RL.
    
    \item \textbf{AD-VAT}~\cite{zhong2019ad}: Introduces an asymmetric dueling mechanism and trains an RL-based tracker with a learnable adversarial target to improve robustness.
    
    \item \textbf{AD-VAT+}~\cite{zhong2019ad+}: An enhanced version of AD-VAT that incorporates a two-stage training scheme, aiming to improve performance in cluttered and obstacle-rich environments.
    
    \item \textbf{TS}~\cite{zhong2023rspt}: A teacher-student framework that extends behavior cloning by leveraging a pose-based teacher to provide real-time supervision for the vision-based student policy during interaction.
    
    \item \textbf{EVT}~\citep{zhong2024empowering}: An offline RL-based framework designed for dynamic target following, which integrates vision foundation models to enhance perception and robustness.

    \item \textbf{IBVS}~\citep{gupta2016novel}: A model-based method that takes the target bounding box as input and applies a Kalman filter-based visual servoing algorithm to follow the target.

    \item \textbf{PoliFormer}~\citep{zeng2024poliformer}: A reinforcement learning-based navigation framework that explicitly encodes target bounding boxes into the observation space to enhance tracking accuracy.
    
    \item \textbf{Uni-NaVid}~\citep{zhang2024uni}: A unified vision-language-action (VLA) model designed for general navigation tasks, including human following.
\end{itemize}

\subsection{Experiment Results}
\textbf{Single Human Tracking Evaluation}
The single human tracking task spans five distinct environments mentioned above, covering a wide range of variations in lighting conditions, viewpoints, and scene layouts. As shown in Table~\ref{tab:eccv_table1}, TrackVLA achieves state-of-the-art performance across all five environments and successfully passes all test cases. Notably,  TrackVLA is trained without any data from this simulator, highlighting its strong generalization under a \textbf{zero-shot} transfer setting.
\begin{table}[htbp]
    \centering
    
    \begin{tabular}{lcccccc}
        \toprule
        Methods & SimpleRoom & Parking Lot & UrbanCity & UrbanRoad & Snow Village & Mean \\
        \midrule
        DiMP & \textbf{500}/\textbf{1.00} & 327/0.48 & 401/0.66 & 308/0.33 & 301/0.43 & 367/0.58 \\
        SARL & \textbf{500}/\textbf{1.00} & 301/0.22 & 471/0.86 & 378/0.48 & 318/0.31 & 394/0.57 \\
        AD-VAT & \textbf{500}/\textbf{1.00} & 302/0.20 & 484/0.88 & 429/0.60 & 364/0.44 & 416/0.62 \\
        AD-VAT+ & \textbf{500}/\textbf{1.00} & 439/0.60 & 497/0.94 & 471/0.94 & 365/0.44 & 454/0.76 \\
        TS  & \textbf{500}/\textbf{1.00} & 472/0.89 & 496/0.94 & 480/0.84 & 424/0.63 & 474/0.86 \\
        RSPT & \textbf{500}/\textbf{1.00} & 480/0.80 & \textbf{500}/\textbf{1.00} & \textbf{500}/\textbf{1.00} & 410/0.80 & 478/0.92 \\
        EVT & \textbf{500}/\textbf{1.00} & 484/0.92 & \textbf{500}/\textbf{1.00} & 496/0.96 & 471/0.87 & 490/0.95 \\
        \rowcolor{gray!20}
        Ours & \textbf{500}/\textbf{1.00} & \textbf{500}/\textbf{1.00} & \textbf{500}/\textbf{1.00} & \textbf{500}/\textbf{1.00} & \textbf{500}/\textbf{1.00} & \textbf{500}/\textbf{1.00} \\
        \bottomrule
    \end{tabular}
    \caption{\textbf{Quantitative results compared with baselines in unseen environments.} The two metrics of each cell represent the Average Episode Length (EL) and Success Rate (SR).}
    \label{tab:eccv_table1}
\end{table}


\textbf{Distraction Robustness Evaluation}
In this experiment, distractors with appearances identical to the target are introduced, requiring the agent to consistently track the first observed target. TrackVLA addresses this challenge using instructions such as “Follow the first person you see”. Experimental results in Table~\ref{tab:eccv_bench2} show that TrackVLA achieves state-of-the-art performance across all scenarios, demonstrating its strong capability in understanding and reasoning about human motion.
\begin{table}[htbp]
    \centering
    
    \begin{tabular}{lccc}
        \toprule
        Methods & Parking Lot (2D) & UrbanCity (4D) & ComplexRoom (4D) \\
        \midrule
        DiMP & 271/0.24 & 348/0.32 & 307/0.26 \\
        SARL & 237/0.12 & 221/0.16 & 263/0.15 \\
        AD-VAT & 232/0.13 & 204/0.06 & 223/0.16 \\
        AD-VAT+ & 166/0.08 & 245/0.11 & 262/0.18 \\
        TS & 331/0.39 & 381/0.51 & 401/0.54 \\
        EVT & 425/0.63 & 472/\textbf{0.92} & \textbf{479}/0.88 \\
         \rowcolor{gray!20}
        Ours & \textbf{467}/\textbf{0.90} & \textbf{476}/\textbf{0.92} & \textbf{479}/\textbf{0.91}  \\
        \bottomrule
    \end{tabular}
    \caption{\textbf{Evaluating the distraction robustness in the environment with distractors.} (4D) represents that there are 4 distractors in the environment.}
    \label{tab:eccv_bench2}
\end{table}

\textbf{Unseen Object Generalization Evaluation}
We further evaluate the object-level generalization ability of TrackVLA using the Gym-UnrealCV benchmark. Specifically, in the \textit{SimpleRoom} environment, we test the model’s tracking performance on four unseen animal categories: horse, dog, sheep, and pig. As shown in Table~\ref{tab:eccv_bench3}, TrackVLA successfully tracks all four categories, consistent with its performance on the single-person tracking task. This demonstrates its strong generalization capability to novel object types.
\begin{table}[htbp]
    \centering
    
    \begin{tabular}{lcccc}
        \toprule
        Methods & Horse & Dog & Sheep & Pig \\
        \midrule
        EVT & \textbf{500}/\textbf{1.00} & 469/0.90 & 471/0.93 & 472/0.94 \\
        \rowcolor{gray!20}
        Ours & \textbf{500}/\textbf{1.00} & \textbf{500}/\textbf{1.00} & \textbf{500}/\textbf{1.00} & \textbf{500}/\textbf{1.00} \\
        \bottomrule
    \end{tabular}
    \caption{\textbf{Evaluating the generalization on the unseen category of the target in SimpleRoom.} We directly adopt the agent on the unseen animals: horse, dog, sheep, and pig.}
    \label{tab:eccv_bench3}
\end{table}

\subsection{Visualization of Humanoid Avatar in Gym-UnrealCV}
\begin{figure}[h]
    \centering
    \includegraphics[width=1\linewidth]{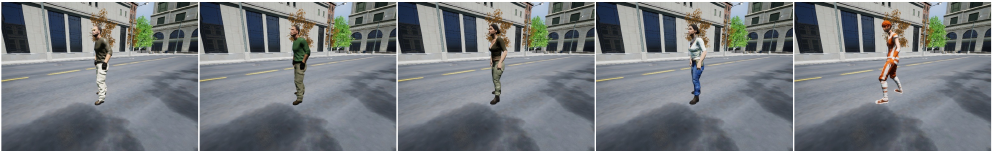}
    \caption{Examples of humanoid avatars used in Gym-UnrealCV.}
    \label{fig:eccv_avatar}
\end{figure}

We showcase several humanoid avatars used in Gym-UnrealCV in Fig.~\ref{fig:eccv_avatar}.

\subsection{Qualitative Results on Gym-UnrealCV}
\begin{figure}[h]
    \centering
    \includegraphics[width=1\linewidth]{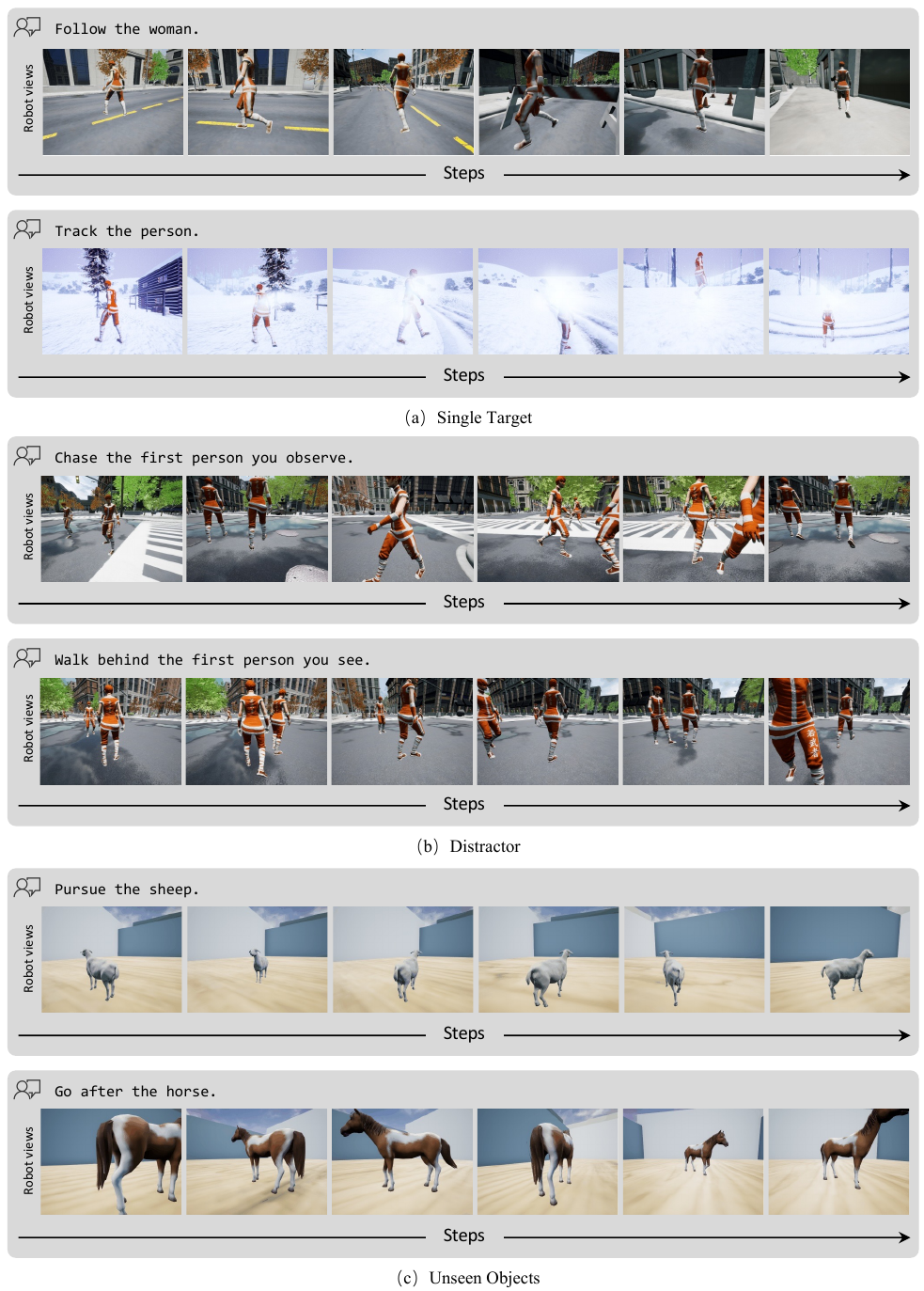}
    \caption{Visualization of TrackVLA on Gym-UnrealCV.}
    \label{fig:eccv_infer}
\end{figure}

We provide visual results of TrackVLA on Gym-UnrealCV, shown in Fig.~\ref{fig:eccv_infer}.

\section{Visual Recognition Experiment}
\subsection{Baselines}
\begin{itemize}[leftmargin=2em]
    \item \textbf{RexSeek}~\cite{jiang2025referring}: A Multimodal Large Language Model (MLLM) designed to detect people or objects in images based on natural language descriptions.
    
    \item \textbf{LISA++}~\cite{yang2023lisa++}: A Multimodal Large Language Model capable of both language understanding and mask generation.
    
    \item \textbf{SoM+GPT-4o}~\cite{yang2023set, openai2024introducing}: A visual prompting method that guides large multimodal models like GPT-4o to perform visual grounding by overlaying segmented image regions with identifiable marks.
\end{itemize}

\subsection{Evaluation}
During testing, each test image contains two \textit{unseen} persons positioned on the left and right sides, and two corresponding descriptions are provided for each person. Given the differing output formats of the evaluated methods, we define task-specific evaluation criteria. RexSeek is an object detection model that outputs bounding boxes; we evaluate its performance by checking whether the predicted box correctly selects the target person. LISA++ is an instance segmentation model that outputs a mask for the target; we assess whether the mask covers the correct individual. The SoM+GPT-4o pipeline first performs image segmentation, then uses SoM to overlay numerical marks on the original image at the location of each segmentation mask. The annotated image is then passed to GPT-4o, which selects the number that best matches the given description. For this method, we evaluate whether the mask corresponding to the selected mark covers the correct individual. As for TrackVLA, which outputs a future trajectory, we determine correctness by checking whether the trajectory direction aligns with the corresponding target person.

\section{More Ablation Study}
\subsection{Action Model Architecture}
The architectures evaluated in this ablation study include Multi-Layer Perceptrons (MLPs) with 3 and 6 layers, respectively, as well as diffusion transformers of varying scales. The hidden state dimensions for the two MLPs are set to 1024 and 4096. The base diffusion transformer is configured with a depth of 12, hidden size of 768, and 12 attention heads, while the small diffusion transformer uses a depth of 6, hidden size of 384, and 4 attention heads.
\begin{table}[h]
    \centering
        \begin{tabular}{lccccc}
        \toprule
        Model & Params. & SR$\uparrow$ & TR$\uparrow$ & CR$\downarrow$ & time(ms) $\downarrow$ \\
        \midrule
        Autoregressive  & 131M  & 42.6  &  56.9    &  11.7    &  460  \\
        MLP (3-Layers)  & 7M  & 45.8  &  59.9    &  10.1    &   \textbf{0.5}   \\
        MLP (6-Layers)  & 89M  & 52.7  &  61.9    &  9.42    &   0.8  \\
        DP-Base & 89M &  17.9  &  33.8    &  27.7    &   65   \\
        Ours-Small &  13M    & 49.8 &   60.2   &  6.67   &  8    \\
        \rowcolor{gray!20}
        Ours-Base & 89M    & \textbf{57.6} &  \textbf{63.2}   &  \textbf{5.80}    & 13    \\
        \bottomrule
        \end{tabular}
    \caption{Comparison of different action models.}
    \label{tab:appendix_action}
\end{table}

\subsection{History Window Length}
Incorporating historical observations helps the model better infer the target’s motion pattern and relative position. Here, we investigate how varying the length of the history observation window $L_{his}$ affects model performance. Table~\ref{tab:ablation_window} shows that removing history observations leads to a significant performance drop. We empirically select 32 as the optimal window length.
\begin{table}[h]
\centering
\begin{tabular}{lccc}
\toprule
$L_{his}$ & SR$\uparrow$ & TR$\uparrow$ & CR$\downarrow$ \\
\midrule
0  & 29.9     &  49.6    &  6.94     \\
\rowcolor{gray!20}
32 & \textbf{57.6}  &  63.2    &   \textbf{5.80}    \\
64 & 56.5     &  \textbf{63.3}    &   6.49    \\
\bottomrule
\end{tabular}
\caption{Comparison of different history window lengths.}
\label{tab:ablation_window}
\end{table}

\subsection{Future Trajectory Horizon}
TrackVLA predicts a future trajectory consisting of $L_{traj}$ waypoints. In Table~\ref{tab:ablation_horizon}, we investigate the impact of varying the number of predicted waypoints on overall performance. Experimental results show that using 10 waypoints yields the best performance.
\begin{table}[h]
\centering
\begin{tabular}{lcccc}
    \toprule
    $L_{traj}$ & SR$\uparrow$ & TR$\uparrow$ & CR$\downarrow$ \\
    \midrule
    1 & 44.3 & 60.6 & 14.4 \\
    \rowcolor{gray!20}
    10 & \textbf{57.6} & \textbf{63.2} & \textbf{5.80}\\
    20 & 51.3 & 60.2 & 7.54 \\
    \bottomrule    
\end{tabular}
\caption{Comparison of different predicted waypoint lengths.}
\label{tab:ablation_horizon}
\end{table}

\subsection{Human Recognition Dataset}
Furthermore, we investigate the impact of different types of human recognition data on the model’s recognition capability. Specifically, we categorize the data into three types: Single Human, Multiple Human, and Same Human, corresponding to images containing one person, 2–3 different individuals, and two identical individuals, respectively. For each category, we construct dedicated human recognition datasets and evaluate the model’s recognition performance under each data setting. Table~\ref{tab:ablation_human} presents the model’s recognition performance under different types of human recognition data. The experimental results demonstrate that the inclusion of each type of human recognition data leads to improved model recognition performance.

In addition, we conduct another analysis to evaluate the impact of removing random backgrounds by replacing all the human recognition data with a plain white background. As presented in Table~\ref{tab:ablation_human}, removing the random background leads to a notable performance drop.
\begin{table}[h]
\centering
\begin{tabular}{cccccc}
\toprule
\makecell{Single\\Human} & \makecell{Multiple\\Human} & \makecell{Same\\Human} & \makecell{Random\\Background} & ACC$\uparrow$ & ACC Drop \\
\midrule
\xmark & \xmark & \xmark & \xmark & 62.0 & 22.9$\%\downarrow$\\
\cmark & \xmark & \xmark & \cmark & 72.3 & 10.4$\%\downarrow$\\
\cmark & \cmark & \xmark & \cmark & 76.7 & 4.60$\%\downarrow$\\
\cmark & \cmark & \cmark & \xmark & 67.4 & 16.2$\%\downarrow$\\
\rowcolor{gray!20}
\cmark & \cmark & \cmark & \cmark & \textbf{80.7} & - \\
\bottomrule
\end{tabular}
\caption{Comparison of different human recognition data.}
\label{tab:ablation_human}
\end{table}

\section{Real-world Deployment}
\begin{figure}[h]
    \centering
    \includegraphics{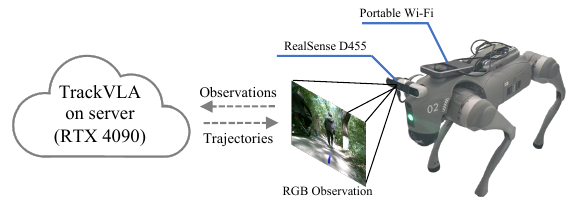}
    \caption{\textbf{Real-world system architecture.} TrackVLA is deployed on a remote server, and the robot communicates with it via the Internet.}
    \label{fig:robot}
\end{figure}
\subsection{Robot Platform}
We provide a visualization of our robotic platform in Fig.~\ref{fig:robot}. The platform is based on the Unitree GO2 quadruped robot, equipped with an Intel RealSense D455 camera. In our work, only RGB frames with a resolution of 640$\times$480 are utilized, under a horizontal field of view (HFOV) of 90$^{\circ}$. Additionally, a portable Wi-Fi is mounted on the back of the robot to enable communication with the remote server through the Internet.

\subsection{Real-world System Architecture}
TrackVLA is deployed on a remote server equipped with an NVIDIA RTX 4090 GPU. During tracking, the server receives the instructions and images captured by the Intel RealSense D455 camera via the Internet. To ensure efficient communication, the images are compressed before transmission. After processing the incoming images, the model performs inference and predicts the future trajectory, which is then transmitted to the quadruped robot for execution. Upon receiving the predicted trajectory, the robot employs a pure pursuit algorithm, combined with its pose information, to perform closed-loop control of its linear and angular velocities, enabling it to follow the trajectory accurately. Additionally, the robot leverages LiDAR point cloud data and implements an elastic band algorithm to achieve obstacle avoidance.

\section{Real-world Experiments}
To further evaluate the tracking capability of TrackVLA, we conducted extensive real-world experiments comparing a quadruped robot powered by TrackVLA with a leading commercial tracking drone (DJI Flip). We tested the following three levels of tracking scenarios with increasing difficulty, each repeated 10 times:

\begin{itemize}[leftmargin=2em]
\item \textit{Easy}: tracking in open outdoor environments without obstacles;
\item \textit{Medium}: tracking in complex environments with occlusions such as walls;
\item \textit{Hard}: tracking a target moving at high speed.
\end{itemize}

\begin{figure}[h]
    \centering
    \includegraphics[width=1\linewidth]{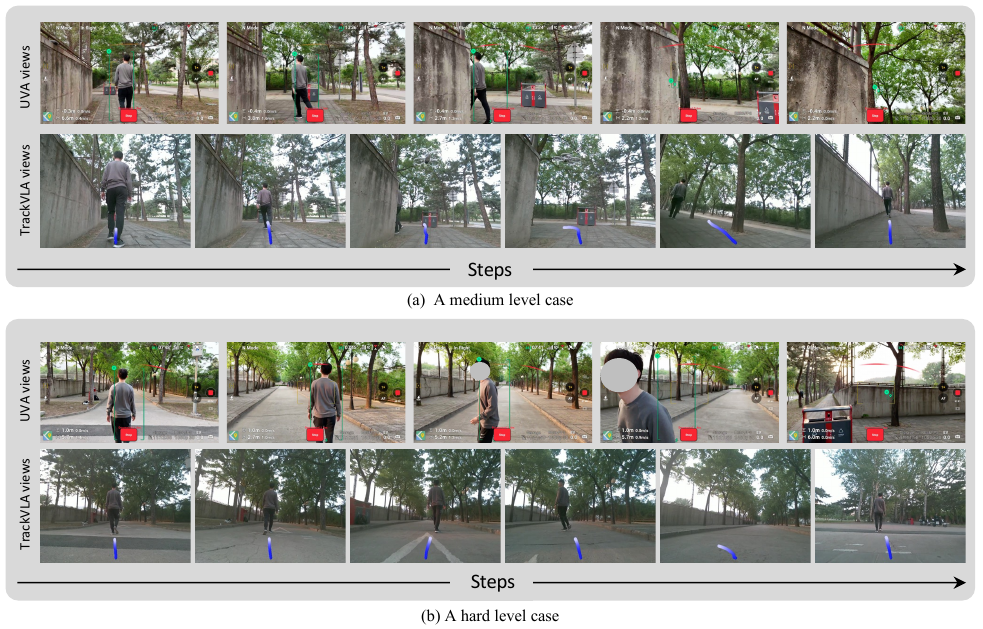}
    \caption{\textbf{Visualization of the real-world experiments.} TrackVLA demonstrates robust tracking performance under challenging conditions such as occlusions and fast target motion, outperforming existing commercial tracking drones.}
    \label{fig:dji}
\end{figure}

The results are shown in Table~\ref{tab:dji}. Both TrackVLA and DJI Flip achieved a 100\% success rate in the \textit{Easy} setting. However, as task difficulty increased, the performance of DJI Flip dropped significantly, falling well below that of TrackVLA. Figure~\ref{fig:dji} further illustrates several representative cases where TrackVLA succeeded while DJI Flip failed. Additional details of the real-world experiments are provided in the supplementary video.

\begin{table}[h]
\centering
\begin{tabular}{lccc}
\toprule
Method & Easy & Medium & Hard \\
\midrule
DJI Flip & \textbf{100\%} & 70\% & 50\% \\
TrackVLA & \textbf{100\%} & \textbf{90\%} & \textbf{70\%} \\
\bottomrule
\end{tabular}
\caption{\textbf{Real-world tracking experiments.} We compare TrackVLA with the commercial tracking drone.}
\label{tab:dji}
\end{table}

\end{document}